%% file: iclr2023_conference_tinypaper.tex
\documentclass{article} 
\usepackage{iclr2023_conference_tinypaper,times}
\usepackage{listings, listings-rust}
\usepackage{graphicx}
\usepackage{multirow}

\input{math_commands.tex}

\usepackage{hyperref}
\usepackage{url}

\title{Discrete Natural Evolution Strategies}

\author{Ahmad Ayaz Amin \\
Department of Computer Science\\
Toronto Metropolitan University\\
Toronto, Canada \\
\texttt{ahmadayaz.amin@torontomu.ca}
}

\iclrfinalcopy 
\begin{document}

\maketitle

\begin{abstract}
Natural evolution strategies are a class of approximate-gradient black-box optimizers that have been successfully used for continuous parameter spaces. In this paper, we derive NES algorithms for discrete parameter spaces and demonstrate their effectiveness in tasks involving discrete parameters.
\end{abstract}

\section{Introduction}
Stochastic gradient descent exploits gradient information to smartly steer parameters towards an optimal solution. However, not all objectives have an easy-to-compute gradient through differentiation, particularly models with discrete parameters. Natural Evolution Strategies (NES) \citep{wierstra2011natural} are a class of black-box optimizers that alleviate the need to compute derivatives, while still retaining gradient-based parameter updates.

At a high level, NES algorithms estimate a gradient by Monte Carlo sampling parameters from a probability distribution known as the search distribution. The sampled parameters are evaluated by a fitness function $f(x_{k})$ and then weighted by the score of the search distribution. The expectation of the scores is the approximate gradient of the search distribution:

\[
    \nabla_{\theta} J(\theta) \approx \frac{1}{\lambda} \sum_{k=1}^{\lambda} f(x_k) \nabla_{\theta} \log \pi(x_k|\theta)
\]

With regular search gradients, vanilla stochastic gradient descent $\theta \leftarrow \theta + \eta \nabla_{\theta} J(\theta)$ updates the parameters of the search distribution. While very effective for complex tasks \citep{salimans2017evolution}, natural evolution strategies employs the \textit{natural} gradient to its advantage, which involves multiplying the gradient with the inverse Fisher information matrix:
\[\theta \leftarrow \theta + \eta \textbf{F}^{-1} \nabla_{\theta} J(\theta)\]

The FIM is a measure of the amount of information that a random variable $x$ carries about the parameters of the distribution from which it was sampled from; it is the variance of the score of the probability distribution.

Using the natural gradient is advantageous, but its costly computational requirements necessitates approximations like Adam \citep{kingma2017adam, JMLR:v21:17-678} for large parameter spaces. Thankfully, not only is the FIM for discrete parameter spaces rather easy to compute, its effects are implicit in regular search gradients, making explicit computation redundant (see Appendix A.3 for details). Nonetheless, we provide the NES updates with the explicit FIM for reference (see Appendix A.1).

The reason for this is that the natural gradient automatically adjust the variance of the search distribution to accelerate convergence. With discrete search gradients, the entropy (the discrete counterpart of variance) readily changes by virtue of updating its parameters after each update.

NES is very similar to variational optimization \citep{staines2012variational} as both estimate the gradient as a Monte Carlo expectation. The major difference lies in the gradient of the search distribution: VO uses the gradient of the probability, while NES uses the score. This subtle difference results in VO computing a lower bound on the true objective, while NES computes an exact gradient in the limit of infinite samples.

\section{Experiment and Results}
A program induction problem is solved using sketching \citep{10.5555/1714168} to demonstrate discrete NES as a practical programming aid. In sketching, a high-level program is specified containing "holes", or incomplete parts, that are completed by a solver. NES is used to complete these holes.

The program is a very simple if-else piecewise function whose discrete parts can be solved easily using brute force. Therefore, holes for coefficients are introduced into the program to make it intractable for brute force computation. These continuous holes are optimized using Gaussian NES \citep{wierstra2011natural}. Discrete holes for operators (e.g. + and *) are solved using discrete NES.

A ground-truth program is created in order to generate input-output pairs (specification) as well as to compare the induced program with.

The sketch is trained to minimize the MSE between the specification output and the inferred output for 10,000 iterations with a learning rate of 0.1. A baseline sketch using VO to solve for the discrete holes is trained for comparison. The training graph is shown below (lower is better):

\begin{figure}[h]
    \centering
    \fbox{\includegraphics[scale=0.35]{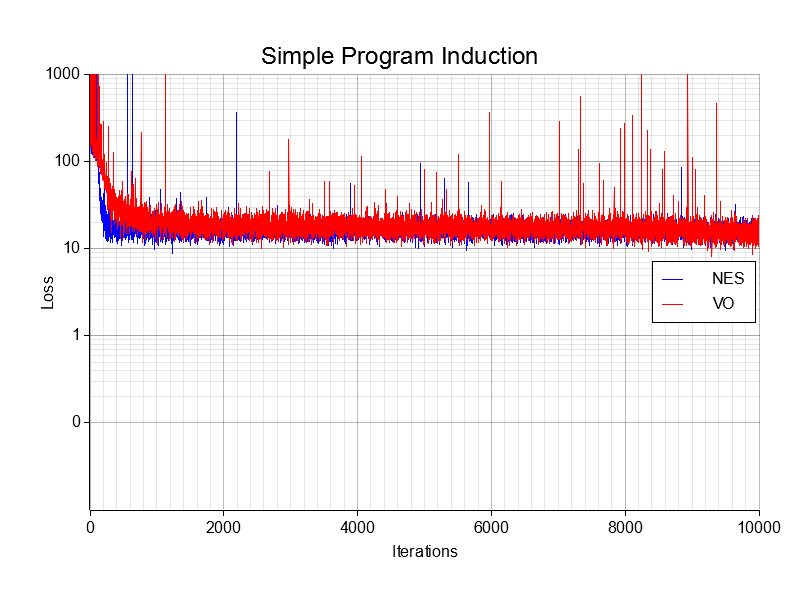}}
\end{figure}

The final outputs of the NES program given the specification inputs [1.0, 2.0, 4.0, 5.0] are [3.9859228, 7.9718456, 15.943691, 19.929615], which are very close to the true outputs [2.1, 4.2, 16.8, 21.0]. The VO sketch had similar results, with its outputs being [2.1309583, 4.2619166, 16.501104, 20.62638]. However, the VO optimizer had notable spikes in the training graph, indicating instability. The low training loss, faster convergence and better stability than VO shows practical utility of discrete NES as a solver for discrete parameter spaces.

\section{Conclusion and Discussion}
In this paper, the natural evolution strategies algorithms for discrete search distributions is derived, allowing optimization of discrete parameters using standard gradient descent optimizers. Like its continuous counterpart, discrete NES demonstrates practical performance on problems.

One quality of discrete NES demonstrated in this paper is that it can effectively work in tandem with continuous NES. However, the experiment is very basic, so it doesn't demonstrate how well it performs on real-life problems with many parameters.

One future avenue is to apply discrete NES to more rigorous problem sets, such as policies with discrete parameters for reinforcement learning \citep{silver2019fewshot}.

\subsubsection*{URM Statement}
The authors acknowledge that at least one key author of this work meets the URM criteria of ICLR 2024 Tiny Papers Track.

\bibliography{iclr2023_conference_tinypaper}
\bibliographystyle{iclr2023_conference_tinypaper}

\newpage

\appendix
\section{Appendix}

\subsection{Derivation of Discrete NES}
\subsubsection{Bernoulli NES}
The log probability mass function $\log \pi(x|\theta)$ of the Bernoulli distribution is $\log \pi (x|\theta) = x\log (\theta)+(1-x)\log (1 - \theta)$ for parameters $\theta \in (0,1)$ and observations $x \in \{0, 1\}$. The score $\nabla_{\theta} \log \pi (x|\theta)$ is thus $\frac{x - \theta}{\theta(1-\theta)}$. The Fisher information of the Bernoulli distribution is $\textbf{F} = \frac{1}{\theta(1-\theta)}$, or the natural gradient of the Bernoulli distribution is:

\begin{equation}
    \nabla_{\theta} \tilde{J}(\theta) \approx \frac{1}{\lambda} \sum_{k=1}^{\lambda} f(x_k) (x_k - \theta)
\end{equation}

\subsubsection{Categorical NES}
Let $\log \pi (x|\theta)$ be the log probability mass function of the categorical distribution: 

\begin{equation}
    \log \pi (x|\theta) = \sum_{k=1}^{K} x_k \log \theta_k
\end{equation}

Then, the partial derivatives are:

\begin{equation}
    \frac{\partial}{\partial \theta_k} \log \pi (x|\theta) = \frac{x_k}{\theta_k}
\end{equation}

Which gives us the following Fisher score:
\begin{equation}
    g(x, \theta) = \begin{bmatrix}
         \frac{x_{1}}{\theta_{1}} \\
         \vdots \\
         \frac{x_K}{\theta_K}
    \end{bmatrix}
\end{equation}

The product of the Fisher score and its transpose is:

\begin{equation}
    g(x,\theta)g(x,\theta)^T = \begin{bmatrix}
         (\frac{x_{1}}{\theta_{1}})^2 & \frac{x_{1}x_{2}}{\theta_{1}\theta_{2}} & \hdots & \frac{x_{1}x_{K}}{\theta_{1}\theta_{K}} \\
         \vdots & \vdots & \hdots & \vdots \\
         (\frac{x_Kx_1}{\theta_K\theta_{1}})^2 & \frac{x_{K}x_{2}}{\theta_{K}\theta_{2}} & \hdots & (\frac{x_{K}}{\theta_{K}})^2
    \end{bmatrix} = \begin{bmatrix}
        g_{11} & g_{12} &\hdots g_{1K} \\
        \vdots & \vdots &\hdots \vdots \\
        g_{K1} & g_{K2} &\hdots g_{KK} \\
    \end{bmatrix}
\end{equation}

For $g_{kk}$, $\E_x[g_{kk}] = \E_x[(\frac{x_k}{\theta_k})^2] = \frac{1}{\theta_k}$. For $g_{ij}$, $i \neq j$, $\E_x[g_{ij}] = \E_x[\frac{x_ix_j}{\theta_i\theta_j}] = 0$. Therefore, the Fisher information matrix for the categorical distribution is $\textbf{F} = \operatorname{diag}\{\frac{1}{\theta_1}, ..., \frac{1}{\theta_K}\}$ \citep{tomczak2012}. Similarly, the inverse FIM is $\textbf{F}^{-1} = \operatorname{diag}\{\theta_1, ..., \theta_K\}$, or the probability vector of the search distribution itself.

The score of the categorical distribution $\nabla_{\theta} \log \pi (x_k|\theta)$ is as follows:

\begin{equation}
    \nabla_{\theta} \log \pi (x_k|\theta) = \begin{cases}
        1 - \pi (x_k|\theta_i) & \mbox{if $k = i$} \\
        - \pi (x_k|\theta_i) & \mbox {if $k \neq i$}
    \end{cases}
\end{equation}

Putting everything together, the natural gradient of the categorical search distribution can be calculated as such:

\begin{equation}
    \nabla_{\theta} \tilde{J}(\theta) \approx \frac{1}{\lambda} \operatorname{diag}\{\theta_1, ..., \theta_K\} \sum_{k=1}^{\lambda} \left[ f(x_k) \begin{cases}
        1 - \pi (x_k|\theta_i) & \mbox{if $k = i$} \\
        - \pi (x_k|\theta_i) & \mbox {if $k \neq i$}
    \end{cases} \right]
\end{equation}

\newpage

\subsection{Program Induction}
The input-output specification was generated using the following Rust program:

\begin{lstlisting}[language=Rust, style=boxed]
fn prog_true(x: f32) -> f32
{
  if x > 3.5
  {
    return 4.2 * x;
  }

  return x * 2.1;
}
\end{lstlisting}

\hspace{10pt}

The program sketch that was optimized has the same general layout:
\begin{lstlisting}[language=Rust, style=boxed]
fn prog_sketch(x: f32) -> f32
{
  if x [COND] [REAL]
  {
    return [REAL] [OP] x;
  }

  return [OP] * [REAL];
}
\end{lstlisting}

The [COND] and [OP] fields indicate conditionals (e.g. ==) and operations (e.g. +) respectively. The [REAL] fields indicate real-valued numbers. The [COND] and [OP] fields are optimized by discrete NES, while the [REAL] fields are optimized by Gaussian NES. At each iteration, the parameters in the fields are sampled from the search distributions and then inserted into the program to evaluate the specification inputs.

\hspace{10pt}

The program resulting from the sketch after 10,000 iterations is as follows:
\begin{lstlisting}[language=Rust, style=boxed]
fn prog_output(x: f32) -> f32
{
  if x < -1.5677981
  {
    return 1.1321394 * x;
  }

  return x * 3.9859228;
}
\end{lstlisting}

With the exception of the order of the conditions, the learned program is functionally similar in both its realization and its outputs to the ground-truth program, indicated by the coefficients and the operators in the return statements. This demonstrates that categorical NES has practical performance, and in turn potential for construction of more expressive models.

\newpage

\subsection{Ablation Study}
In this section, the effect of the FIM in training performance is investigated to evaluate whether FIM is necessary for better convergence behaviour. 

The experiment is a slightly more sophisticated program induction problem involving two inputs (i.e. a multi-variable piecewise function) in order to demonstrate performance under increased complexity. The same sketch is used for both the explicit NES as well as the search gradients (no FIM) optimizer.

Similar to the main experiment, discrete search distributions are used for operators and continuous search distributions are used for real-valued constants. The experiment is performed with a series of different learning rates ranging from 0.1 to 0.001 to investigate the behaviour of the two optimizers. Each run is performed for 10,000 iterations.

The inputs for the program are the pairs [(5.8, 2.5), (5.0, 6.2), (7.4, 6.1), (5.5, 9.4)]. The outputs produced by the ground-truth program are [14.1, -4.677419, 20.9, -5.287234].

\hspace{10pt}

\begin{tabular}{|p{2cm}||p{2cm}|p{3cm}|p{2cm}|p{3cm}|}
     \hline
     \multicolumn{5}{|c|}{Effect of FIM on Convergence After 10,000 Iterations} \\
     \hline
     Learning Rate & NES Loss & NES Outputs & SG Loss & SG Outputs \\
     \hline

     $0.1$ & $36.94874$ & [$19.529205$, $-3.3582695$, $10.21171$, $-6.8166084$] & $36.948715$ & [$19.529203$, $-3.3576767$, $10.211709$, $-6.8160696$] \\
     \hline
     $0.05$ & $36.949665$ & [$19.515165$, $-3.3503892$, $10.204369$, $-6.8094444$] & $36.949516$ & [$19.515043$, $-3.3440251$, $10.204305$, $-6.803659$] \\
     \hline
     $0.01$ & $38.834015$ & [$19.653248$, $-5.3543134$, $10.27657$, $-8.631194$] & $37.132545$ & [$19.62102$, $-3.9629102$, $10.259719$, $-7.3662815$] \\
     \hline
     $0.005$ & $53.801556$ & [$15.546269$, $0.0117374435$, $8.129055$, $0.0070379255$] & $39.12318$ & [$19.64944$, $-5.504375$, $10.27458$, $-8.767613$] \\
     \hline
     $0.001$ & $170.9544$ & [$0.047796037$, $0.022356212$, $0.0153531805$, $0.013405079$] & $133.90349$ & [$3.8961499$, $1.3543346$, $2.0372744$, $0.98261297$] \\
     
     \hline 
\end{tabular}

\hspace{10pt}

Strangely enough, it seems as though the FIM has a negative impact on the discrete NES, as indicated by the gradually increasing loss and diverging outputs with respect to learning rate. In contrast, search gradients maintains lower and more stable loss values under these changes. 

This behaviour is not entirely unexpected, as the FIM is only applicable to continuous search distributions where the variance of the distribution requires continuous adjustment with respect to the optima \citep{wierstra2011natural}. 

In the case of a Gaussian search distribution, the variance parameter $\sigma ^ 2$ needs to be large when the parameters are far from the optimal solution, and small when the parameters are within a close radius. This is to prevent potential training instabilities and faster convergence.

In the case of the categorical search distribution, the entropy (the categorical equivalent of variance) readily changes with gradient updates. In other words, the behaviour of the natural gradient is implicitly built into the search gradient update. Thus, adding the extra FIM is not only redundant and computationally wasteful, it is actually detrimental to model performance. This is further emphasized by the fact that the inverse FIM for the categorical distribution is the probability vector of the distribution itself.

\end{document}

%% file: math_commands.tex
\usepackage{amsmath,amsfonts,bm}

\def\eqref#1{equation~\ref{#1}}

\def\1{\bm{1}}

\DeclareMathAlphabet{\mathsfit}{\encodingdefault}{\sfdefault}{m}{sl}
\SetMathAlphabet{\mathsfit}{bold}{\encodingdefault}{\sfdefault}{bx}{n}

\newcommand{\E}{\mathbb{E}}

